\documentclass{article}
\usepackage{graphicx}
\usepackage{setspace}
\usepackage{xcolor}
\usepackage{amsmath}
\usepackage{hyperref}
\usepackage{float}
\usepackage{authblk}
\usepackage{caption}
\usepackage{times}          \usepackage{mathptmx}
\usepackage[numbers]{natbib}
\usepackage[a4paper, margin=1in]{geometry}
\usepackage{tcolorbox}
\usepackage{todonotes}
\usepackage{booktabs}
\usepackage{array}
\usepackage{longtable}
\usepackage{tikz}
\usetikzlibrary{positioning,arrows.meta,calc}
\usepackage{url}
\hypersetup{
  colorlinks=true,
  linkcolor=black,
  citecolor=black,
  filecolor=black,
  urlcolor=black
}

\tcbset{colback=cyan!10!white, colframe=cyan!50!black}
\date{}

\title{The challenge of uncertainty quantification of large language models in medicine}
\author[1]{Zahra Atf}
\author[2]{Seyed Amir Ahmad Safavi-Naini (MD)}
\author[1]{Peter R. Lewis}
\author[7]{Aref Mahjoubfar (MD)}
\author[2]{Nariman Naderi (MD)}
\author[5,6]{Thomas R. Savage (MD)}
\author[2,3,4]{Ali Soroush (MD)}

\affil[1]{Faculty of Business and Information Technology, Ontario Tech University, Oshawa, Canada}
\affil[2]{Division of Data-Driven and Digital Medicine (D3M), Icahn School of Medicine at Mount Sinai, New York, NY 10029, United States}
\affil[3]{The Charles Bronfman Institute of Personalized Medicine, Icahn School of Medicine at Mount Sinai, New York, NY 10029, United States}
\affil[4]{Henry D. Janowitz Division of Gastroenterology, Icahn School of Medicine at Mount Sinai, New York, NY 10029, United States}
\affil[5]{Department of Medicine, Stanford University, Stanford, United States}
\affil[6]{Division of Hospital Medicine, Stanford University, Stanford, CA 94304, United States}
\affil[7]{School of Medicine, Iran University of Medical Sciences, Tehran, Iran}
\begin{document}
\maketitle  
\begin{abstract}
\setlength{\baselineskip}{0.9\baselineskip}
This study investigates uncertainty quantification in large language models (LLMs) for medical applications, emphasizing both technical innovations and philosophical implications. As LLMs become integral to clinical decision-making, accurately communicating uncertainty is crucial for ensuring reliable, safe, and ethical AI-assisted healthcare. Our research frames uncertainty not as a mere impediment but as an inherent aspect of knowledge that invites a dynamic and reflective approach to AI design. By integrating advanced probabilistic methods—such as Bayesian inference, deep ensembles, and Monte Carlo dropout—with linguistic analysis that computes predictive and semantic entropy, we propose a comprehensive framework that differentiates and manages both epistemic and aleatoric uncertainties. The framework further incorporates surrogate modeling to circumvent the limitations of proprietary APIs, multi-source data integration for enhanced context sensitivity, and dynamic calibration techniques via continual and meta-learning strategies. In addition, explainability is embedded through the development of uncertainty maps and composite confidence metrics, which aim to bolster user trust and enhance clinical interpretability. By aligning uncertainty metrics with real-world clinical risk factors, our framework supports decision-making processes that are transparent, ethically responsible, and aligned with the principles of both Responsible and Reflective AI.
Philosophically, our approach challenges the conventional pursuit of absolute predictability by advocating for the acceptance of controlled ambiguity—a shift that encourages the development of AI systems that are not only technically robust but also reflective of the inherent provisionality of medical knowledge.
\end{abstract}

\textbf{Keywords:} Uncertainty Quantification, Large Language Models, Medical AI, Epistemic Uncertainty, Aleatoric Uncertainty.

\section{Introduction}
Large Language Models (LLMs) have rapidly gained prominence as powerful tools capable of understanding and generating human-like text. Their adoption in high-impact domains such as finance, law, and healthcare underscores the urgent need for these systems to reliably communicate uncertainty. When LLMs are expected to perform tasks where errors carry significant consequences, such as clinical decision support, a key challenge arises: they must not only produce accurate answers but also signal low confidence when uncertainty is high \cite{Sheng2024}. This challenge, often conceptualized as selective classification or classification with a reject option, has driven research across machine learning, learning theory, and natural language processing \cite{Feng2021}\cite{jones2021}\cite{El-Yaniv2010}\cite{Bartlett2008}\cite{KamathJ2020}\cite{liang2023}\cite{xiong2024}. Traditional techniques typically rely on the model’s softmax probabilities or internal representations to approximate and convey uncertainty \cite{jones2021}\cite{liang2023}.

Despite these foundational efforts, there remains a pressing need to systematically examine and refine uncertainty management strategies for LLMs. This necessity is most apparent in healthcare applications, where reliable and safe AI-assisted decisions are paramount. Medical queries demand up-to-date, evidence-based answers, but the inherent stochastic nature of text generation in LLMs can yield inconsistent outputs. Slight variations in model settings or prompt formulations can change the model’s predictions, posing significant risks when patient safety is on the line. Moreover, biases embedded in training data, along with potential knowledge gaps, can further erode the trustworthiness of an LLM in clinical settings.

The ability to accurately quantify and effectively communicate uncertainty serves as a cornerstone for safe deployment of LLMs in healthcare \cite{Savage2024}. Clinicians, researchers, and healthcare administrators rely on these models to provide high-confidence recommendations or to indicate when further expert input is necessary. Yet, persistent challenges remain, including limited guidance on curating and validating healthcare data sets and the ambiguity of source attribution, given the vast corpus of untraceable text data leveraged by models like ChatGPT \cite{Lin2024}. These issues underscore the importance of establishing robust frameworks that not only ensure transparency and accountability but also allow diverse user groups to interpret and act on model outputs with an appropriate level of caution.

This article aims to fill these gaps by comprehensively reviewing current methods for uncertainty quantification and management in LLMs, with a particular emphasis on healthcare. Through this focus, we seek to highlight both the technical and ethical dimensions of adopting such models in clinical practice, where any shortcomings can have life-altering consequences. By framing the discussion around uncertainty, we point to the pivotal need for explainability frameworks that can bolster trust in these models. Ultimately, the insights gained from this review will guide future research and policy efforts to ensure that the potential of LLMs in healthcare is realized responsibly—balancing the pursuit of innovation with the imperative of patient safety and equitable care \cite{shrivastava2023llamasknowgptsdont}\cite{donakanti2024reimaginingselfadaptationagelarge}.
our work is in the topic of the meaning of uncertainly and reviewing to all vertion of this lecture and this reviwing, 
to confidence of this research of the lecture. 

\section{Understanding Uncertainty and Risk}
Uncertainty is a fundamental concept across diverse disciplines such as artificial intelligence (AI), philosophy, statistics, and the social sciences, significantly influencing both theoretical discourse and practical applications. In many of these fields, understanding uncertainty is closely tied to assessing and managing risk, as both concepts play a pivotal role in decision-making processes under incomplete or ambiguous information \cite{Wu2020}\cite{kochenderfer2022}.
Risk assessment and predictability have conceptual overlap but serve distinct purposes within decision-making models. For example, while risk focuses on known probabilities, uncertainty pertains to unknown or unpredictable outcomes \cite{aven2019risk}. Therefore, designing systems with clear, explainable, and transparent methods for defining risk and uncertainty can enhance both user trust and system reliability \cite{rudin2019interpretable}.

\subsection{Philosophical and Ethical Perspectives on Uncertainty}
In philosophy, the concept of uncertainty refers to the inability to attain absolute knowledge or accurately predict future events. This notion is particularly relevant within the domains of epistemology and the philosophy of science \cite{Dlugatch2024}. From a philosophical standpoint, uncertainty challenges the scope of human knowledge, raising epistemological inquiries about what can be known with certainty \cite{gorichanaz2020epistemology}. Similarly, fuzzy logic—a computational framework rooted in epistemic and ontological discussions on vagueness and indeterminacy—acknowledges that truth values can exist in degrees rather than binary forms \cite{bede2019fuzzy}. By extending these reflections into ethics, uncertainty will also raise issues of responsibility, fairness, and transparency in AI. The notion of overlapping concepts emerges when multiple terms—such as trust, reliability, and fairness—are used interchangeably or intersect, rendering it difficult to delineate their boundaries clearly \cite{Freyer2024}\cite{Hallowell2022}\cite{Hullermeier2021}.

Although explainability is often considered a crucial factor in fostering trust in AI, research has demonstrated that explanations do not always lead to increased trust. In some cases, users may perceive explanations as deceptive, overly technical, or even unnecessary, depending on their prior beliefs and level of expertise. This phenomenon has been explored in prior work, where it was shown that explanations can sometimes reinforce pre-existing skepticism rather than alleviate it \cite{Atf2023HumanCI}. Furthermore, recent studies highlight that trust in AI systems is shaped by various socio-cognitive factors, including users’ subjective perceptions, cognitive biases, and prior experiences with automated decision-making \cite{mehrabi2021bias}. These findings align with broader philosophical perspectives emphasizing the need for critical approaches to managing uncertainty \cite{Hallowell2022}\cite{Dlugatch2024}. In clinical contexts, for example, empirical knowledge must be integrated with scientific data and clinical judgment to navigate uncertainty, yet absolute certainty remains unattainable \cite{Muller2024}. This underscores the complexity of trust-building in AI, reinforcing the notion that explainability alone is insufficient without considering the broader psychological, ethical, and contextual dimensions of human-AI interaction thinking in decision-making processes \cite{Hallowell2022}.

\subsection{Epistemic and Aleatoric Uncertainty}
In AI and machine learning, uncertainty is frequently categorized into two principal types: epistemic uncertainty, which reflects a lack of knowledge about the model or system, and aleatoric uncertainty, representing the inherent randomness in the data \cite{kendall2017}. Uncertainty is also prominent in the social sciences and behavioral studies, where human emotions, intentions, and social dynamics introduce significant ambiguity. However, it is more accurate to attribute this ambiguity to the limitations of the measurement tools rather than the phenomena being measured.

In the realm of biostatistics, which is often how physicians conceptualize uncertainty, aleatoric uncertainty, more commonly referred to as statistical uncertainty, is particularly pronounced in scenarios such as cancer metastasis, where the dissemination of malignant cells occurs in a stochastic manner \cite{CASTANEDA202217}\cite{naderi2025}; disease remission, where recovery or relapse follows unpredictable patterns \cite{Berkhof2009}; and multiple sclerosis (MS) flare-ups, which are often triggered by random factors such as environmental influences or immune system fluctuations \cite{Lebel2023}. These examples underscore the intrinsic randomness of medical phenomena, presenting significant challenges for models that must integrate data-driven uncertainty.
Moreover, the dynamic nature of patient data—such as evolving symptoms, variable treatment responses, and newly emerging diagnostic insights—further complicates uncertainty quantification. To mitigate these challenges, incorporating real-time data updates and integrating information from multiple sources can enhance the adaptability of predictive models, including LLMs, to continuously evolving clinical settings \cite{Edoh2024}.

A key challenge in understanding uncertainty lies in distinguishing between epistemic uncertainty and ontological uncertainty \cite{Wu2020}\cite{Nwebonyi2024}\cite{Dlugatch2024}. Epistemic uncertainty pertains to limitations in knowledge and the availability of information, while ontological uncertainty is rooted in the intrinsic unpredictability of reality, where certain phenomena remain inherently unknowable \cite{Dlugatch2024}.
Ontological uncertainty is particularly evident in various aspects of medical science, where intrinsic complexities challenge deterministic modeling. For example:

\begin{itemize}
    \item \textbf{Chronic Illness:} The progression of diseases such as diabetes and autoimmune disorders is influenced by a multitude of interacting factors, including diet and stress, making precise deterministic predictions unattainable \cite{ANDREASSI200935}\cite{Lebel2023}.
    \item \textbf{Psychosocial Complexity:} A patient’s cultural background and socioeconomic status can unpredictably affect treatment adherence and health outcomes \cite{Williams2010}\cite{Devenish2017}\cite{Vaughn2009}.
    \item \textbf{Biological Chaos:} Nonlinear interactions within biological systems, such as immune responses or epigenetic regulation, contribute to unpredictable variations in medical conditions.
    \item \textbf{Emergent Phenomena:} Large-scale effects like herd immunity \cite{Khanjanianpak2022} or antibiotic resistance \cite{Sharif2021} arise from complex micro-level interactions \cite{Anderson2018}, further complicating predictability in medical science.
\end{itemize}

Proper management of these uncertainties is imperative to ensure the reliability of AI systems. Overlapping concepts like trustworthiness and explainability are crucial in this context. For instance, systems that provide uncertainty estimates can enhance users' trust by rendering predictions more transparent and comprehensible \cite{BARREDOARRIETA202082}.

\section{Drivers of Response Uncertainty}

When we narrow our focus to the realm of LLMs, especially in the medical domain, additional challenges arise due to the nature of language generation and the complexity of clinical data. Studies have shown that the stochastic behavior inherent in LLMs leads to variability not only in the text output but also in the confidence levels reported, necessitating novel quantification methods that combine linguistic analysis with probabilistic modeling \cite{Wang2020}.
In particular, integrating uncertainty metrics with medical context—such as correlating predictive entropy with clinical risk factors—could enhance the reliability of AI-assisted diagnostic systems. This is critical when decisions depend on the model's ability to indicate low confidence for ambiguous or contradictory information \cite{Hosseini2023}.

Uncertainty in AI-generated responses has many sources, including data inputs and outputs, the intricacies of model architecture, user interactions, and the contextual milieu in which the AI operates. Each of these elements contributes uniquely to the aggregate uncertainty of the system, necessitating a holistic analytical approach to enhance the reliability, interpretability, and trustworthiness of AI models—particularly in high-stakes domains such as healthcare or financial decision-making, where the precision of outcomes is paramount \cite{abdar2021review}.

The management of uncertainty is pivotal across many fields, exerting a profound influence on both theoretical constructs and real-world applications. In the realm of AI, it assumes a central role in ensuring that models can adeptly accommodate the complexities inherent in real-world scenarios.

In the social sciences, uncertainty serves as a fundamental factor shaping human behavior and societal trends. Researchers have delved into how individuals make decisions within uncertain environments, illuminating psychological, economic, and policy-related challenges \cite{kahneman2011thinking}.

\begin{figure}[H]
\centering
\includegraphics[trim={0 0 0 0}, clip, width=1\textwidth]{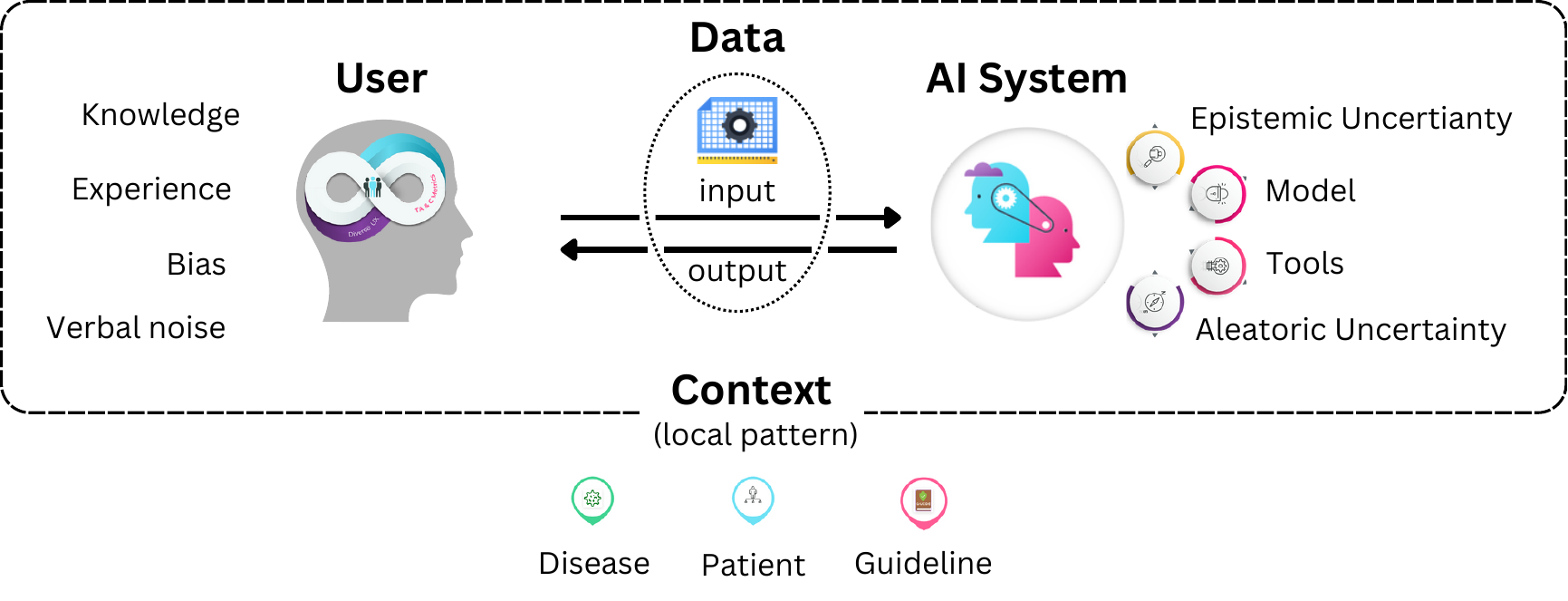} 

\caption{\textbf{Various Drivers of Response Uncertainty.} This diagram illustrates how user knowledge and biases, data quality (inputs and outputs), the AI system’s architecture, and context (such as local disease patterns or practice routines) interact to influence the level of uncertainty in model-generated responses. Each element contributes uniquely to overall system reliability, interpretability, and trustworthiness.}
\label{fig:all_drivers}
\end{figure}

An integrated perspective on uncertainty across different disciplines offers crucial insights for AI research. Robust AI systems hinge on effective uncertainty management to predict outcomes with greater accuracy under conditions of ambiguity \cite{kochenderfer2022decision}. Moreover, statistical methodologies provide invaluable tools for quantifying uncertainty, thereby enhancing decision-making processes in situations where information is incomplete or unreliable \cite{gelman2014bayesian}. A deeper comprehension of decision-making patterns, such as risk aversion \cite{kahneman1979prospect}, heuristic reliance \cite{Meghan2024}, and adaptive strategies \cite{Collins2024}, facilitates the development of more resilient and adaptive systems and policies, aiding both individuals and organizations in navigating uncertainty more effectively. \textbf{Figure \ref{fig:all_drivers}} provides a visual overview of how user factors, data inputs and outputs, the AI model itself, and the broader clinical or operational context collectively drive uncertainty in an LLM’s responses.

By systematically accounting for uncertainty in data, model behaviors, user interactions, and contextual variables, AI systems can achieve heightened transparency, reliability, and user trust, thereby fostering a more harmonious alignment between automated decision-making processes and human expectations.

\subsection{Data (Input and Output)}
\begin{figure}[H]
\centering
\includegraphics[trim={0 0 0 0}, clip, width=1\textwidth]{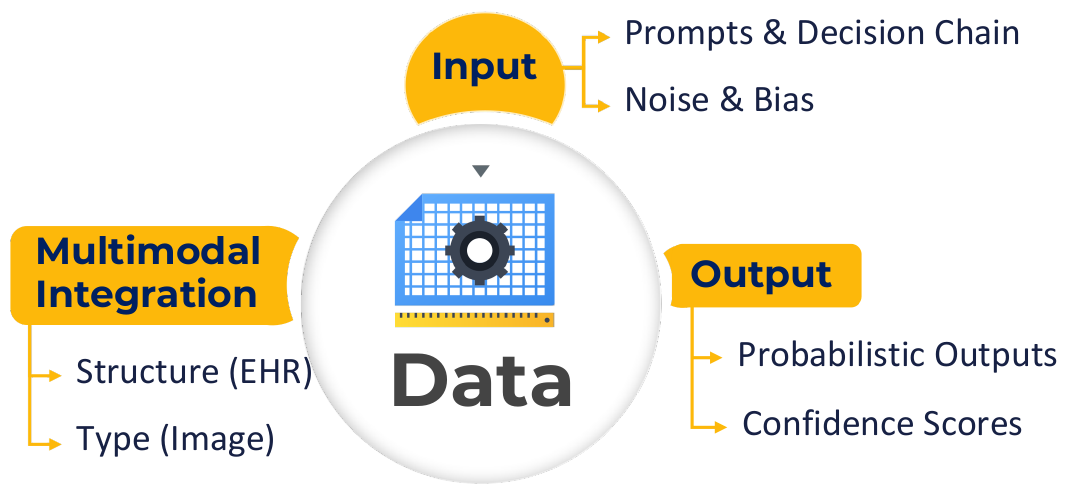} 

\caption{\textbf{Data as the Driver of Uncertainty: input quality, output quality, and integration.} This diagram highlights the key factors influencing data uncertainty in AI systems. It shows how multimodal data integration, including structured data (e.g., Electronic Health Records) and unstructured data (e.g., images), along with biases introduced by system designed (system prompts and the decision chain) and user (noise, bias and mistakes in user prompts), impact the quality of inputs and the probabilistic nature of model outputs.}
\label{fig:data-uncertainty}
\end{figure}

Studies on uncertainty quantification in LLMs for medical applications utilize a diverse range of data inputs, including both structured clinical records and unstructured text, such as clinical notes, research articles, and patient narratives. The quality, completeness, and nature of these inputs significantly influence the uncertainty of LLM predictions. For instance, noisy, biased, or incomplete clinical datasets can lead to variability in LLM outputs, while ambiguous language often results in multiple plausible interpretations that diminish the confidence of diagnostic or prognostic suggestions\cite{qi2020data}\cite{zhou2020uncertainty}.
LLMs are designed to produce outputs that may include probabilistic diagnostic recommendations, risk assessments, or explanatory text. These outputs are often augmented with confidence scores or uncertainty estimates to help clinicians interpret the results effectively. However, ambiguous phrasing and language nuances in the model's responses add an extra layer of complexity to uncertainty quantification, which is particularly critical in high-stakes medical decision-making contexts \cite{harsha2024quantifying}.
Multi-source data integration is also a key component of advancing LLM performance and reducing uncertainty. In the medical domain, this entails combining inputs from various sources such as electronic health records, imaging reports, genetic data, and even patient-generated content from wearable devices. This integrative approach helps ensure that the models capture a comprehensive view of patient health, but it also requires sophisticated aggregation algorithms and uncertainty estimation techniques, especially when operating under privacy-preserving paradigms like federated learning\cite{Khawaled2024}\cite{ChenEtal2025}\cite{kairouz2021advances}.
Evaluations on both synthetic and real-world clinical datasets are crucial for assessing model robustness and for ensuring reliable performance under domain shifts. The inconsistency between training data distributions and real-world scenarios often leads to unpredictable model performance on unseen cases, highlighting the need for continuous monitoring and refinement of uncertainty quantification mechanisms\cite{mehrabi2021bias}\cite{Fazekas2023}.
By addressing these challenges, the research on uncertainty quantification in LLMs aims to improve the transparency, reliability, and clinical applicability of AI-driven recommendations in medicine. \textbf{Figure \ref{fig:data-uncertainty} }illustrates the various data-related factors influencing uncertainty and uncertainty estimation, focusing on the quality of inputs, the resulting outputs, and the integration of multimodal data sources.

\subsection{Model}
\begin{figure}[H]
\centering
\includegraphics[trim={0 50 0 100}, clip, width=1\textwidth]{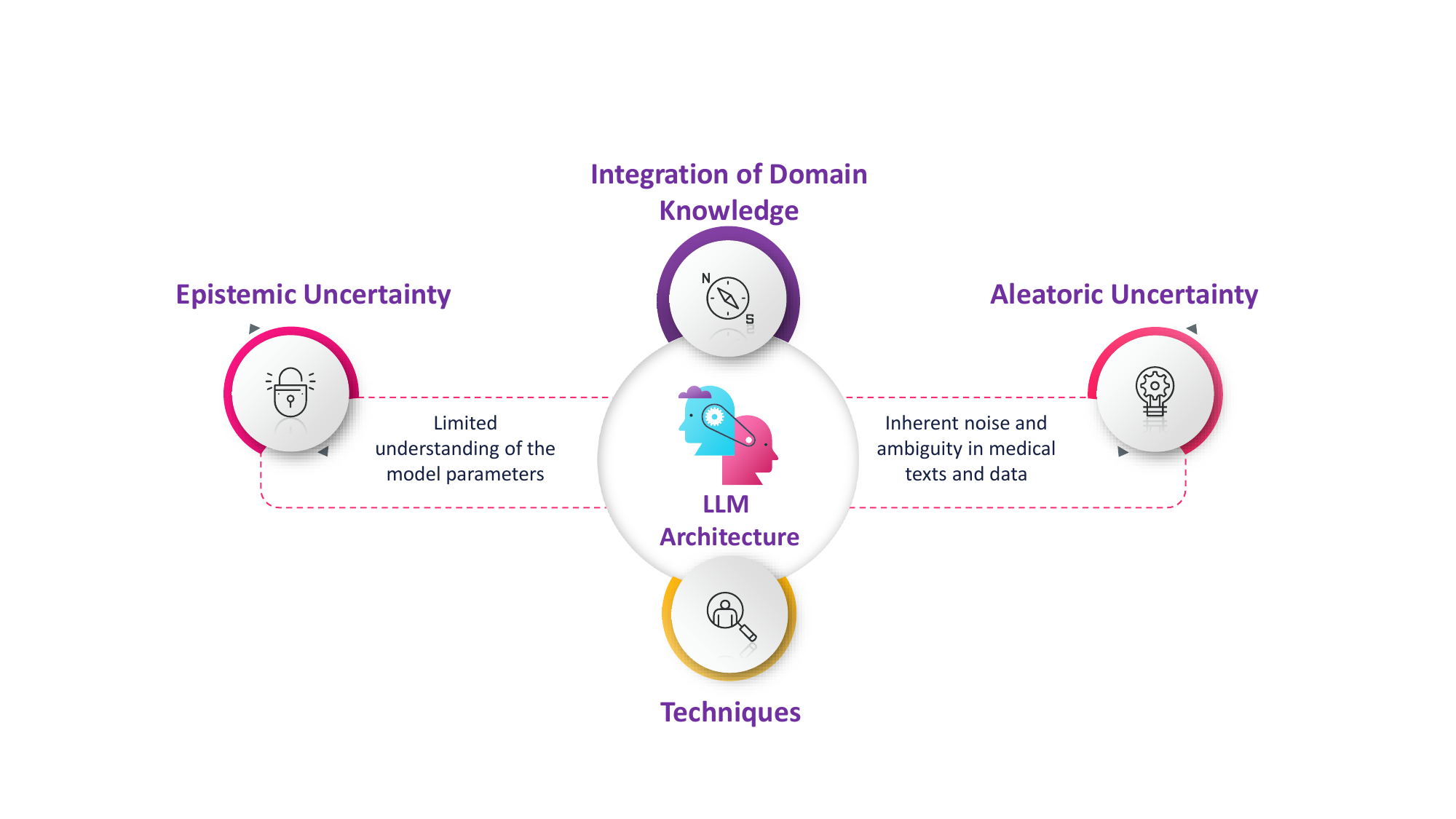} 

\caption{\textbf{Key Components of LLM Architecture for Uncertainty Quantification in Medicine.} This diagram shows the primary factors influencing uncertainty in large language models (LLMs) used in healthcare. It highlights the relationship between epistemic uncertainty (due to limited understanding of model parameters) and aleatoric uncertainty (caused by inherent noise and ambiguity in medical data). The integration of domain-specific knowledge and advanced techniques in LLM architecture aims to manage and quantify these uncertainties, improving model reliability and interpretability in medical applications.}
\label{fig:LLM-uncertainty}
\end{figure}

Recent studies on uncertainty quantification have extended to LLMs, particularly in the context of medical applications. Advanced architectures for LLMs are now being designed to address both epistemic uncertainty and aleatoric uncertainty \cite{gawlikowski2021survey}.
While earlier work in domains like medical imaging employed convolutional neural networks (CNNs) combined with Transformers or U-shaped architectures to harness both local and global information \cite{Xiao2022}, similar hybrid approaches are being explored for LLMs. These approaches integrate both traditional natural language processing (NLP) techniques and modern deep learning architectures to manage uncertainty in clinical narratives, electronic health records, and medical literature.
Techniques such as Bayesian neural networks and dropout regularization have been adapted for LLMs to provide reliable uncertainty estimates \cite{kendall2017uncertainties}. Moreover, recent advances in deep evidential learning have shown promise in LLMs; for instance, deep evidential regression methods can learn evidential distributions over outputs, capturing both types of uncertainties efficiently without the need for computationally expensive sampling methods or ensembles \cite{Liu2021}. This is critical for ensuring that models can provide transparent and trustworthy recommendations in healthcare contexts.
Furthermore, integrating domain-specific medical knowledge—including clinical terminologies and diagnostic guidelines—into the architecture of LLMs enhances both interpretability and performance. Techniques such as surrogate modeling and the incorporation of probabilistic loss functions contribute to this goal, collectively boosting the model’s reliability in supporting complex medical decision-making processes\cite{Salvador2023}\cite{Dimitriou2023}\cite{Chen2025}.
\textbf{Figure \ref{fig:LLM-uncertainty}} outlines the key components of LLM architecture posing challenge to uncertainty quantification, specifically emphasizing how epistemic and aleatoric uncertainties interact with the techniques and integration of domain knowledge to influence model predictions.

\subsection{User}
\begin{figure}[H]
\centering
\includegraphics[trim={0 0 0 0}, clip, width=1\textwidth]{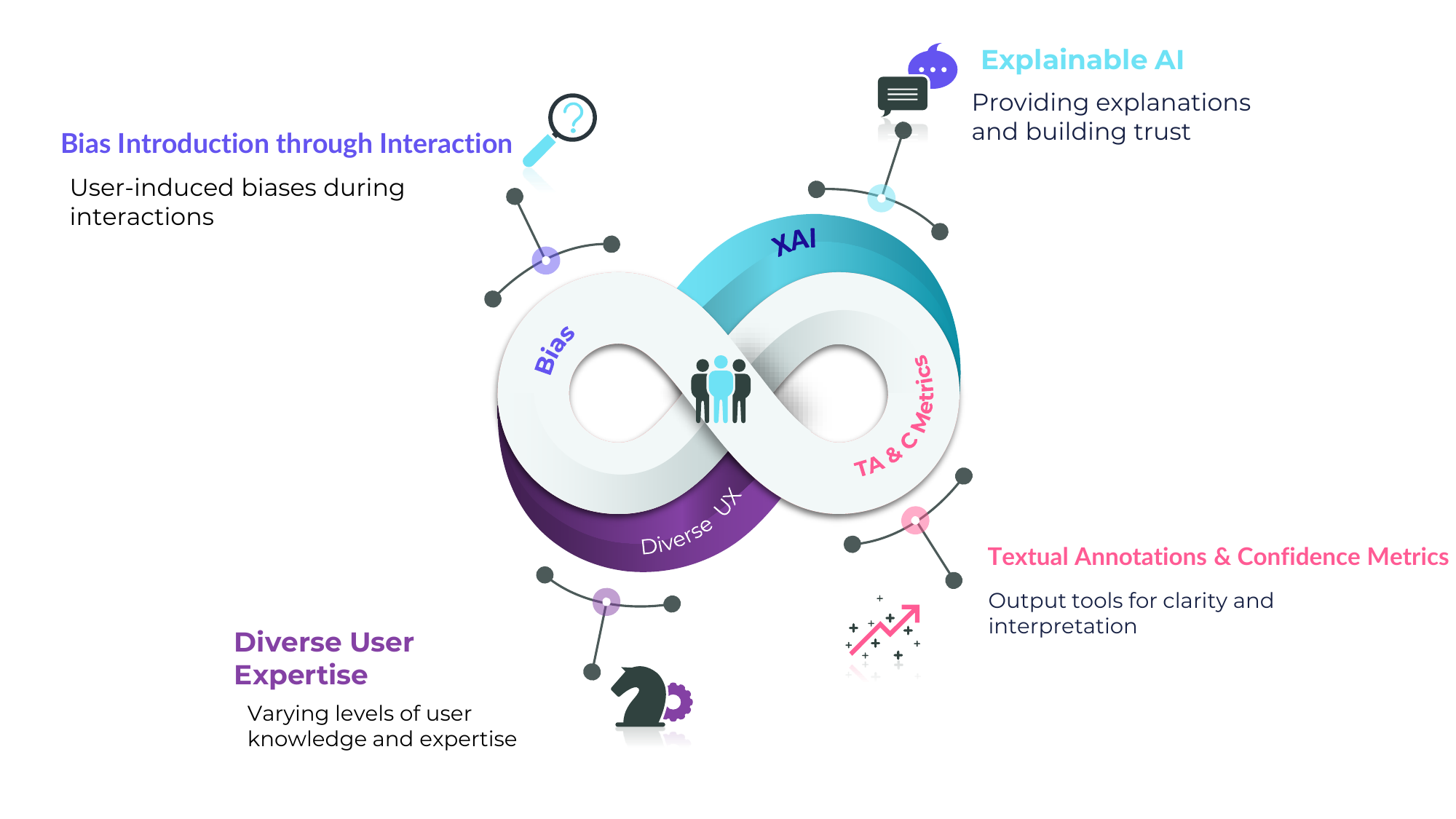} 

\caption{\textbf{User-Centric Factors Impacting Uncertainty Quantification in LLMs for medicine.} This diagram illustrates the various factors that influence how users interact with large LLMs in medical contexts. It shows the impact of user-induced biases, diverse levels of user expertise, and the role of explainable AI (XAI) in providing clarity through textual annotations and confidence metrics. These elements collectively shape how uncertainty is perceived and managed in AI-assisted decision-making.}
\label{fig:user-uncertainty}
\end{figure}

Uncertainty-aware LLMs in medicine are primarily employed by clinicians and researchers to support high-stakes tasks such as diagnostic interpretation, treatment planning, and patient monitoring\cite{Nurhidayah2023}\cite{lanfredi2024enhancingchestxraydatasets}. In medical contexts, these LLMs are designed not only to deliver accurate and reliable information but also to address user concerns by incorporating explainable AI (XAI) techniques, structured explanations, and clear uncertainty metrics\cite{Bhutto2024}\cite{Campos2023}. Tools such as textual uncertainty annotations, confidence indicators, and probabilistic output statements help enhance clinician trust and engagement by providing interpretable and actionable insights\cite{Huang2025}\cite{Shen2023}\cite{Darley2023}.
User interactions introduce an additional layer of complexity regarding uncertainty. Clinicians and medical professionals come from diverse backgrounds and possess varying expectations and levels of expertise. These differences can influence how they interpret and interact with the outputs of LLMs, potentially leading to misaligned expectations. Such misalignment may result in misunderstandings, reduced confidence in the system, or even suboptimal use of the LLM’s capabilities\cite{yin2019understanding}\cite{yang2020unremarkable}. Additionally, user interactions can inadvertently introduce biases or errors, further complicating the uncertainty landscape in medical decision-making\cite{eiband2021impact}.
By providing clear explanations and interpretable outputs, these models enable clinicians to understand the underlying reasoning behind diagnostic suggestions or treatment recommendations, thereby reducing uncertainty and enhancing trust\cite{adadi2018peeking}\cite{arrieta2020explainable}. Moreover, involving end users in the design and iterative refinement of LLM systems—through participatory design and ongoing feedback mechanisms—can further align the models with real-world clinical needs, minimizing uncertainties stemming from misaligned expectations \cite{muller2021shifting}. Collaborative approaches, such as human-in-the-loop systems and robust feedback loops, are essential to supporting clinicians as they navigate ambiguous or uncertain scenarios, ultimately fostering seamless integration of LLMs into clinical workflows \cite{Bora2024}.
\textbf{Figure \ref{fig:user-uncertainty}} highlights the user-centric factors that influence uncertainty quantification in LLMs, including user-induced biases, diverse user expertise, and the role of XAI and output annotations in enhancing model interpretability and trust

\subsection{Context}
Recent advancements in uncertainty-aware AI have been applied across numerous medical domains, from clinical decision support and patient education to remote health monitoring and clinical trial design\cite{Xiao2022}\cite{Molchanova2025}\cite{Kucukgoz2025}\cite{ChenEtal2025}. For LLMs in medicine, contextual factors are of paramount importance. The same medical query or clinical note may yield divergent responses depending on patient demographics, local disease prevalence, or evolving medical guidelines\cite{situ2021context}. This variability emphasizes the need for LLMs to understand and adapt to changing clinical environments—where contextual uncertainty arises when models lack sufficient information about the broader clinical setting.
A structured approach to understanding and addressing contextual challenges in medicine is the PEAS Framework (Performance, Environment, Actuators, Sensors), which defines the task environment for rational agents. In the medical domain, this framework can be applied as follows:\cite{russell2020artificial}

\textbf{Performance Measure:} Achieving high diagnostic accuracy, optimizing patient prognosis (e.g., survival rates), minimizing treatment-related risks (e.g., side effects), reducing invasiveness, ensuring cost-effectiveness, adhering to clinical guidelines, and maximizing patient satisfaction.

\textbf{Environment:} Functioning within diverse healthcare settings, including hospitals, clinics, telemedicine platforms, electronic health records (EHRs), imaging systems, wearable devices, and direct interactions with patients and healthcare personnel.

\textbf{Actuators:} Producing actionable outputs such as treatment recommendations, clinical alerts (e.g., early sepsis detection), automated documentation, test ordering, and patient communication tools.

\textbf{Sensors:} Gathering data from various sources, including patient history, physical examinations, laboratory test results, imaging data, real-time vital signs (e.g., heart rate), and clinician notes.

Medical task environments possess unique characteristics that influence how uncertainty is managed:

\textbf{Observability:} Medical settings are often partially observable, as clinicians and AI systems rarely have access to complete patient data.

\textbf{Agents:} These environments involve multi-agent interactions, including healthcare providers (e.g., doctors, nurses, specialists), patients, AI systems (e.g., diagnostic tools, electronic records, medical devices), and regulatory bodies (e.g., guidelines, laws, ethical frameworks).

\textbf{Determinism:} Patient responses to treatments are inherently nondeterministic, making outcomes difficult to predict.

\textbf{Temporal Dependency:} Medical decisions often have long-term, sequential consequences rather than being isolated events.

\textbf{Dynamism:} Patient conditions can change rapidly, making the medical environment highly dynamic.

\textbf{Data Representation:} Many medical parameters, such as vital signs (e.g., blood pressure) and lab values (e.g., glucose levels), are continuous variables, whereas diagnostic classification tasks often involve discrete categories (e.g., determining the presence or absence of a disease).

\textbf{Knowledge Gaps:} Medical knowledge remains incomplete, particularly for conditions lacking well-defined guidelines, such as rare genetic disorders.

\begin{figure}[H]
\centering
\includegraphics[trim={0 100 0 100}, clip, width=1\textwidth]{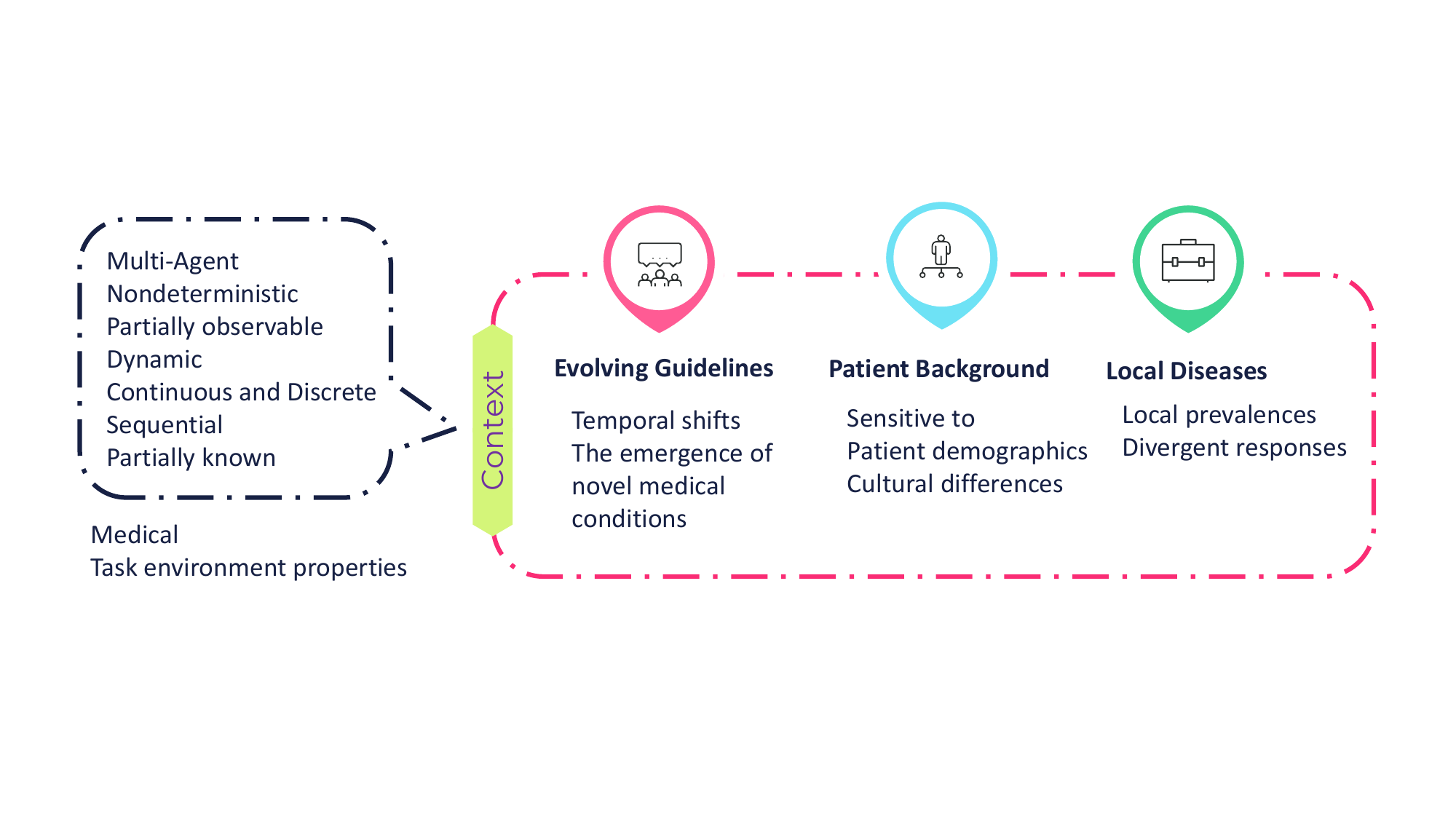} 

\caption{\textbf{Key Contextual Factors Influencing Uncertainty in LLMs for Medical Applications.} This diagram highlights the primary contextual elements that affect uncertainty in medical AI systems. It shows how factors such as evolving medical guidelines, patient demographics, and local disease patterns influence uncertainty in model predictions. These elements are part of the broader medical task environment, which is characterized by dynamic, multi-agent, and partially observable properties.}
\label{fig:context-uncertainty}
\end{figure}

These properties highlight the necessity of integrating context-aware mechanisms into AI models for medical applications, ensuring they effectively manage uncertainty while improving clinical decision-making.
Incorporating contextual awareness into LLMs involves ensuring that these models are sensitive to variables such as patient background, cultural differences, temporal shifts in healthcare practices, and the emergence of novel medical conditions \cite{chen2020context}\cite{Liebermann2024}. Techniques like meta-learning and continual learning have been introduced to improve the models' ability to generalize across diverse medical tasks and adapt seamlessly to new clinical contexts with minimal additional training data \cite{hospedales2022meta}. Such strategies are particularly critical for LLMs, which must deliver accurate and contextually appropriate outputs in dynamic and often uncertain clinical settings, including those encountered in wearable health monitoring and integrated clinical workflows \cite{Yan2023}\cite{Khawaled2024}\cite{Darley2023}.
Moreover, contextual uncertainty not only impacts diagnostic or prognostic outputs but also shapes how LLM outputs are perceived and used by healthcare professionals. For example, in patient education or remote health monitoring, the clarity and contextual relevance of a language model's response can directly affect patient outcomes and adherence to medical advice \cite{lanfredi2024enhancingchestxraydatasets}\cite{Trowman2021}\cite{Karthik2025}. Addressing these challenges requires that LLMs provide not just accurate answers but also interpretable uncertainty metrics that reflect real-world contextual dynamics.
\textbf{Figure \ref{fig:context-uncertainty}} illustrates the key contextual factors that influence uncertainty in LLMs used in medical applications, emphasizing how evolving regional guidelines, patient background, and local diseases impact uncertainty quantification.

By integrating robust contextual awareness and adaptive learning strategies, LLMs can significantly enhance their clinical relevance and reliability, ensuring that uncertainty quantification truly supports informed, context-sensitive decision-making in medicine.

\subsection{Interplay of Components}
\begin{figure}[H]
\centering
\includegraphics[trim={0 10 0 10}, clip, width=1\textwidth]{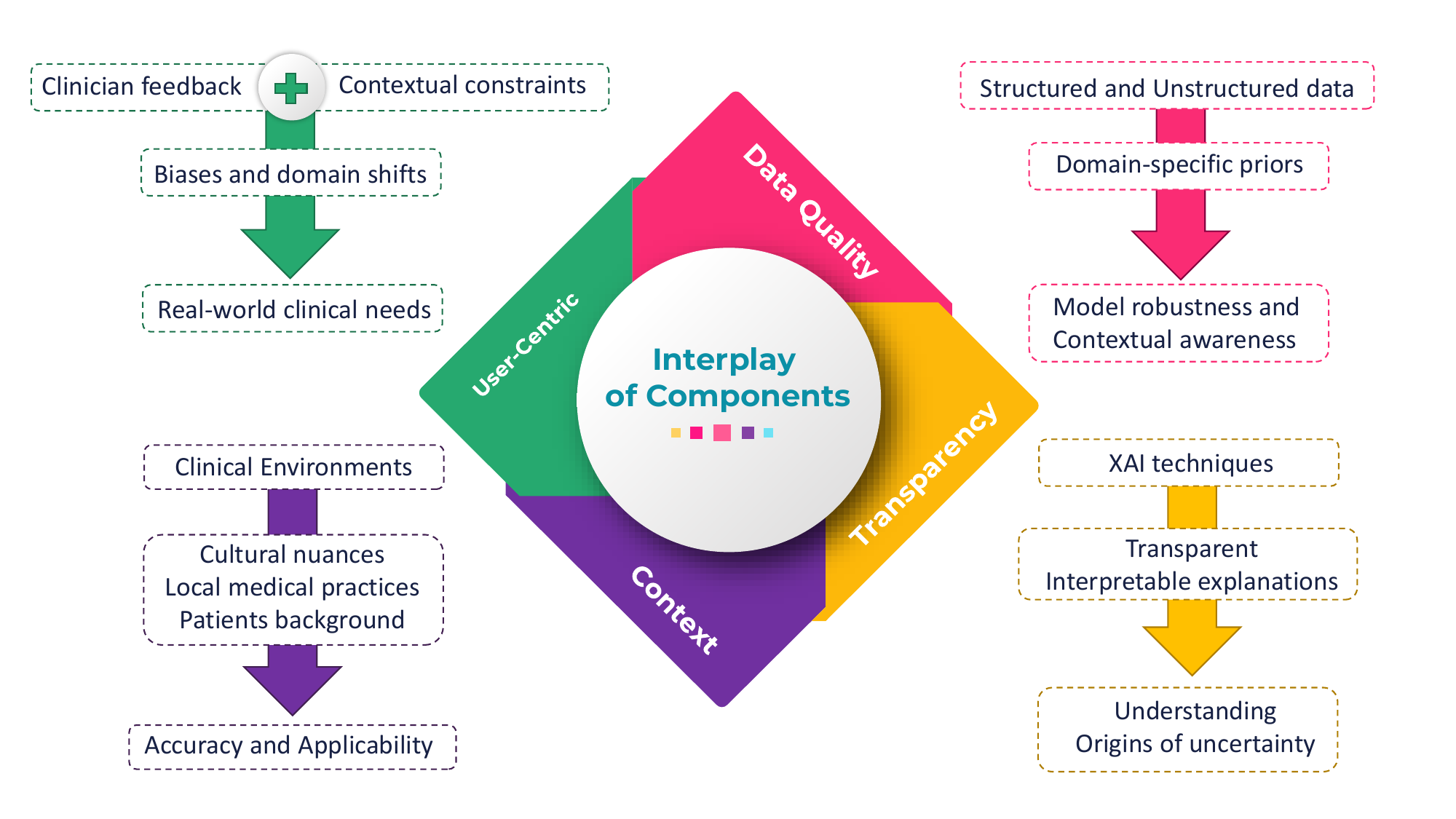} 

\caption{\textbf{Core Components of Uncertainty Management in LLMs: data, transparency, user design, and context.} This diagram highlights the critical components involved in managing uncertainty within LLMs. It demonstrates the relationship between user-centric design, data quality, transparency, and context, as well as the influence of these factors on model performance. The model's robustness, contextual awareness, and the application of explainable AI (XAI) techniques play an essential role in ensuring transparent and interpretable outputs while considering real-world clinical environments and user feedback.}
\label{fig:uncertaintyManagemenet}
\end{figure}

In practice, the interaction between input data, the LLM's architecture, user interpretation, and clinical context significantly compounds overall uncertainty. A model trained on incomplete or biased patient data may offer predictions that overlook critical cultural or context-specific nuances. When clinicians encounter such outputs, there may be misalignments with local medical practices or expectations, which can reduce trust and hinder the model's integration into clinical workflows \cite{Baghirov2024}.
To bridge this gap, XAI methods have become indispensable in the realm of LLMs for medicine. By providing transparent and interpretable explanations of decision-making processes, XAI techniques allow clinicians to understand the origins of uncertainty in model predictions. This transparency not only helps in reconciling model outputs with clinical judgment but also fosters greater trust and usability in high-stakes medical environments\cite{Ong2025}. Robust uncertainty management frameworks, therefore, play a critical role in aligning LLM predictions with actual clinical needs and practices, enhancing both interpretability and real-world applicability\cite{Trowman2021}.
Recent advancements in the deployment of LLMs for medical applications underscore the importance of integrating diverse data sources, domain-specific knowledge, and adaptive modeling techniques to both enhance accuracy and manage uncertainty\cite{Xiao2022}. In the context of LLMs, combining structured and unstructured data (e.g., clinical notes, patient records, and medical literature) with domain-specific priors helps in building models that are not only robust but also contextually aware. This fusion is further enriched through hybrid model architectures that leverage both statistical learning and rule-based domain insights.
Collaborative frameworks that integrate model outputs with clinician feedback and contextual constraints are emerging as key strategies in addressing data quality issues, representation biases, and domain shifts \cite{Koume2025}\cite{Dimitriou2023}. For instance, when an LLM processes ambiguous clinical queries, uncertainty metrics—such as confidence scores or probability distributions over potential interpretations—assist in guiding users through potential diagnostic or therapeutic pathways, especially in rare or complex cases. Data augmentation strategies, guided by these uncertainty measures, along with adaptive training techniques, further contribute to maintaining consistent performance across a variety of clinical scenarios \cite{Bhutto2024}\cite{Khawaled2024}\cite{Chen2025}.
\textbf{Figure \ref{fig:uncertaintyManagemenet} } summarizes the core components of uncertainty management in LLMs, focusing on the interplay between data quality, transparency, context, and user-centric design to improve model reliability and interpretability.

\section{Uncertainty Quantification}

Biostatistical methods for uncertainty quantification and mitigation have long been used in clinical prediction modeling. Classical approaches—such as bootstrapping, cross‐validation, and Bayesian shrinkage—are routinely utilized to address model instability and to provide reliable estimates of prediction error. Bootstrapping, for example, estimates the sampling distribution of regression coefficients and prediction errors, while cross‐validation offers a robust means to assess out-of-sample performance and calibration. In addition, Bayesian shrinkage techniques integrate prior information to reduce overfitting, resulting in more stable confidence intervals and improved risk assessments\cite{Heinze2018}.

Recent advances in hybrid modeling, which combine deep learning techniques with Bayesian inference, have shown promise in reducing the epistemic uncertainty in LLM outputs. Such models not only improve calibration but also provide interpretable uncertainty estimates that can inform clinical decision-making \cite{Liu2021}. Moreover, explainability frameworks that visualize uncertainty (e.g., uncertainty maps) and correlate them with clinical parameters are being developed to bridge the gap between raw statistical measures and the practical needs of healthcare professionals \cite{KamathJ2020}.

Integrating insights from philosophy, AI, and the social sciences allows for the development of more robust frameworks to handle uncertainty across various domains \cite{Williams2010}. These broader definitions of uncertainty help bridge gaps between statistical models and human-centric approaches, ensuring AI systems are not only reliable but also aligned with the complexities of real-world decision-making \cite{arrieta2020explainable}.\\
In clinical practice, various reasoning models used by physicians offer structured frameworks to navigate uncertainty. These models are grounded in key principles such as Heuristic Use (Rule of Thumb) and Occam's Razor, which advocate for simpler explanations when multiple hypotheses are presented \cite{Sterkenburg2025} ; the Pauker-Kassirer model, which formalizes decision-making based on probabilities, utilities and decision thresholds \cite{Pauker1980} ; “ex juvantibus” reasoning, where therapeutic trials inform diagnostic insights \cite{Fahrlander1985} ; and in more general Bayesian reasoning, which updates probabilities based on new evidence. By incorporating such principles, clinicians can systematically approach diagnostic and therapeutic uncertainties, fostering a balance between efficiency and thoroughness in decision-making.
This interdisciplinary perspective demonstrates the importance of combining technical methodologies, philosophical frameworks, and clinical expertise to effectively manage uncertainty. Such integration not only enhances the interpretability of AI models but also aligns them with the nuanced needs of medical practitioners.

Understanding the multifaceted nature of uncertainty in medical decision-making is not merely an academic endeavor; it is a critical necessity for ensuring the reliability and safety of AI-driven systems in healthcare. Addressing challenges, arising from various sources, requires integrating systematic approaches, such as advanced probabilistic models, robust data preprocessing techniques, and uncertainty-aware training frameworks.\\
Moreover, the implications of uncertainty quantification extend far beyond technical aspects, shaping ethical and practical dimensions of AI adoption in medicine. Healthcare professionals need transparent models that not only provide accurate predictions but also explain the confidence levels behind these predictions. Patients, in turn, benefit from informed decision-making that acknowledges inherent risks and promotes a collaborative approach to care. By exploring and addressing uncertainty, we set the stage for building systems that align with the core principles of trust, safety, and fairness in medical applications.\\

\subsection{Methodological Challenges and Emerging Gaps}

The growing adoption of LLMs in diverse fields—particularly in medicine—has highlighted the need for a comprehensive understanding of both uncertainty quantification and explainability in these systems. This review of the literature consolidates insights from multiple studies to examine various types of uncertainty, the methodologies applied to estimate it, the strategies used to clarify model outputs, and the real-world challenges these models encounter. \textbf{Table \ref{table:arpproaches-in-uncertainty-in-med}} offers an in-depth overview of key concepts, technical hurdles, and the effects of uncertainty on trustworthiness within medical applications of LLMs.

\begin{longtable}{|>{\raggedright\arraybackslash}p{0.25\textwidth}|>{\raggedright\arraybackslash}p{0.25\textwidth}|>{\raggedright\arraybackslash}p{0.25\textwidth}|>{\raggedright\arraybackslash}p{0.25\textwidth}|}
\caption{\textbf{Key Themes and Approaches for Managing Uncertainty in Medicine}} 
 \label{table:arpproaches-in-uncertainty-in-med} \\
\hline
\textbf{Theme} & \textbf{Definition of Uncertainty} & \textbf{Challenges in Measuring Uncertainty} & \textbf{Role of Uncertainty in Trustworthiness} \\
\hline
Techniques for uncertainty quantification and improving trust in semi-supervised models. & Explores Monte Carlo Dropout, MSE, and predictive entropy as primary methods for uncertainty quantification\cite{Xiao2022}\cite{Yan2023}.
Highlights improved model trust by reducing uncertainty in semi-supervised learning. & Discusses model calibration and stability\cite{Silk2022}\cite{Jones2025}. Examines Gaussian Process Emulators (GPE) for parameter optimization\cite{Jones2025}. & Highlights the role of reducing uncertainty in increasing reliability in medical imaging models\cite{Xiao2022}\cite{Wickstrøm2021}. \\
\hline
Ensemble methods for addressing domain shifts and uncertainty reduction. & Describes deep ensembles and multi-scale uncertainty analysis as foundational methods for reducing uncertainty\cite{Molchanova2025}\cite{Du2025}. & Examines domain shift challenges using ensemble methods\cite{Du2025}. & Suggests reducing uncertainty to improve prediction accuracy and trust in medical contexts\cite{Du2025}. \\
\hline
Uncertainty visualization for improving prediction clarity and trust. & Introduces dropout techniques and uncertainty maps for clarity in predictions\cite{Campos2023}\cite{Kucukgoz2025}\cite{Huang2025}. & Emphasizes the role of data preprocessing for uncertainty management\cite{Huang2025}. & Improves transparency and trust in sensitive applications through uncertainty visualization\cite{Kucukgoz2025}. \\
\hline
Psychological and user-centric approaches to uncertainty management. & Highlights the psychological effects of uncertainty and its management in model outputs\cite{Nurhidayah2023}\cite{Shen2023}. & Focuses on inadequate data and interpretability challenges\cite{Shen2023}. & Recommends clear explanations and user-centric designs for improved trust in AI\cite{Nurhidayah2023}. \\
\hline
Computational methods for addressing domain shifts and improving trust. & Emphasizes computational issues and data quality in uncertainty quantification\cite{Chen2025}\cite{ChenEtal2025}. & Discusses methods like OOD detection and thresholding for managing domain shifts\cite{Wang2025}. & Links reducing uncertainty to improved trust and explainability in decision-making systems\cite{ChenEtal2025}. \\
\hline
Use of explainability to enhance trust and reliability. & Advocates for transparent uncertainty metrics, trust scores, and visual aids\cite{Haq2025}\cite{Darley2023}\cite{Vreman2022}. & Highlights the role of uncertainty explainability in improving decision reliability\cite{Darley2023}. & Demonstrates how trust increases through explainability in medical applications\cite{Vreman2022}. \\
\hline
Lexicon-based methods for classification and trust improvement. & Uses supervised learning and specialized lexicons to classify and manage uncertainty in outputs\cite{Wiedmann2024}\cite{McGowan2025}. & Examines the limitations of classification methods for uncertainty assessment\cite{Wiedmann2024}. & Proposes leveraging structured vocabularies to improve user trust in predictions\cite{McGowan2025}. \\
\hline
Dynamic feedback systems for adaptive uncertainty management. & Discusses the need for dynamic feedback systems to manage adaptive uncertainties\cite{Wiedmann2024}\cite{McGowan2025}. & Emphasizes the role of continuous model recalibration and iterative learning\cite{Palmer2023}. & Highlights adaptive uncertainty management to build user trust in real-time systems\cite{Mennin2023}. \\
\hline
Bayesian models and maps for transparency and trust. & Highlights Bayesian models and uncertainty maps for better explainability\cite{Kang2024}\cite{Popat2023}. & Examines computational limits in Bayesian uncertainty quantification\cite{Popat2023}. & Shows how transparency in uncertainty boosts user trust in LLMs and enhances model credibility\cite{Kang2024}. \\
\hline
\end{longtable}

In the definition section, approaches such as employing dropout techniques, deep ensembles, multi-scale analyses, and Bayesian models are considered to reduce ambiguity in predictive outputs, thereby enhancing the transparency and reliability of AI-driven systems. These approaches emphasize the importance of providing both visual and textual explanations to better understand the uncertainty in model outputs, demonstrating that reducing this type of uncertainty can play a crucial role in improving the performance of diagnostic and therapeutic systems.\\
On the other hand, the existing challenges in measuring uncertainty include issues related to model calibration, result stability, computational limitations, and the necessity of proper data reprocessing. These issues cause models to become misaligned when encountering shifts in data distributions and real-world conditions. Moreover, enhancing the transparency of outcome reporting and implementing dynamic feedback systems for continuous performance improvement are vital tools for assuring users and medical professionals. Another key takeaway from the table is the significance of integrating technical and human aspects in uncertainty management; while advanced computational techniques are employed to reduce output ambiguities, attention to the psychological aspects and direct communication with users (such as physicians) is equally important. This demonstrates that creating transparency in results and offering comprehensive explanations not only improves the performance of AI models but also increases their acceptance in sensitive medical settings.
\\
\\
\subsection{Approaches for Enhanced Uncertainty Quantification}
Given the inherent complexities of medical data and the critical need for accurate diagnoses, hybrid approaches that integrate diverse datasets with optimized machine learning algorithms have become indispensable. These methods mitigate uncertainty while enhancing the accuracy, transparency, and reliability of medical data analysis. Table 2 succinctly summarizes the methods and techniques employed for uncertainty quantification in medical LLMs, highlighting various applications and corresponding outcomes.\\
A key solution to managing uncertainty involves integrating hybrid models that combine linguistic confidence estimates with numerical surrogate models. This approach compensates for the lack of access to internal probabilities in proprietary APIs like GPT-4\cite{Yan2023}. Surrogate models, such as Llama-2, improve AUROC and overall calibration by providing internal probabilities, which are used alongside token-level probabilities to refine predictions. Additionally, combining semantic entropy with sample consistency methods helps detect incorrect responses more effectively, especially in medical decision-making\cite{Liebermann2024}\cite{Shen2023}. These methods offer a robust means of uncertainty quantification, ensuring that both aleatoric and epistemic uncertainties are properly managed across different tasks.
\\

\begin{longtable}{|p{0.18\textwidth}|p{0.18\textwidth}|p{0.28\textwidth}|p{0.28\textwidth}|}
\caption{\textbf{Overview of Methods and Applications for LLM Uncertainty Quantification.}} \label{table:methods-for-UQ-LLM}
\\
\hline
\textbf{Theme} & \textbf{Topic} & \textbf{Methods and Techniques} & \textbf{Applications and Results} \\ 
\hline
\endfirsthead
\hline
\textbf{Theme} & \textbf{Topic} & \textbf{Methods and Techniques} & \textbf{Applications and Results} \\ 
\hline
\endhead
\hline
\endfoot

\raggedright Focuses on reducing noise and improving uncertainty representation through probabilistic learning. & 
Importance of Uncertainty Assessment & 
\raggedright Use of supervised learning models and Bayesian Neural Networks (BNNs). BNNs employ probabilistic distributions instead of point estimates to better capture uncertainty\cite{Wiedmann2024}\cite{9263102}. & 
Improved predictive accuracy and identification of uncertainty in model outputs. These methods also enhance performance by reducing data noise \cite{Wiedmann2024}\cite{9263102}. \\ 
\hline

Distinguishes between two main uncertainty types, enhancing model interpretability. & 
Bayesian Frameworks and Probabilistic Methods & 
Techniques like Monte Carlo Dropout and Deep Ensembles introduce stochasticity into predictions to calculate uncertainty distributions\cite{9263102}\cite{Sabeti2021}. & 
Accurate distinction between epistemic uncertainty (model-related) and aleatoric uncertainty (data-related), improving human-model collaboration\cite{9263102}\cite{Sabeti2021}. \\ 
\hline

Automates text-based uncertainty detection, improving analysis speed. & 
Automated Text Classification Techniques & 
Use of models such as Lasso and SVM to identify uncertainty-related paragraphs in large texts\cite{Wiedmann2024}. & 
Faster and more accurate detection of uncertainty in texts, with applications in analyzing LLM responses\cite{Wiedmann2024}. \\ 
\hline

Establishes domain-specific vocabularies for handling clinical uncertainty. & 
Uncertainty Management in ICU & 
Thematic analysis and Delphi processes to establish a vocabulary for clinical uncertainty\cite{McGowan2025}. & 
Helps LLMs better understand clinical uncertainty concepts and improves performance in EHR-based predictions\cite{McGowan2025}. \\ 
\hline

Specializes in handling temporal data with high-dimensional uncertainty. & 
Time-Series Predictive Models & 
Use of Dirichlet-multinomial models and MCMC simulations to calculate uncertainty distributions\cite{Bartolucci2021}. & 
More accurate predictions of ICU capacity, mortality rates, and patient durations, while accounting for data uncertainty\cite{Bartolucci2021}. \\ 
\hline

Prioritizes safety by integrating human expertise for high-uncertainty cases. & 
Clinical Decision Support & 
Referral of high-uncertainty cases to human experts using referral mechanisms\cite{9263102}. & 
Enhances human-model collaboration and reduces diagnostic errors in high-risk cases\cite{9263102}. \\ 
\hline

Improves user trust by offering visual explanations. & 
Model Interpretability Techniques & 
Methods like Grad-CAM, LIME, and LRP to improve model transparency\cite{Teng2024}. & 
Increases model transparency, builds user trust, and facilitates clinical decision-making with comprehensible explanations\cite{Teng2024}. \\ 
\hline

Quantifies the effect of uncertainty on model performance. & 
Evaluating the Impact of Uncertainty on Performance & 
Use of Permutation Testing and variance analysis to assess model performance under uncertainty\cite{Xue2024}. & 
Reduces diagnostic errors and improves the interpretability of outputs in high-dimensional and variable data scenarios\cite{Xue2024}. \\ 
\hline

Establishes a structured framework for multi-level uncertainty classification. & 
Multi-Level Frameworks for Uncertainty Evaluation & 
Use of the GRADE framework to classify different levels of uncertainty in models\cite{Rothenberger2025}. & 
Standardizes evaluations and enhances comparability of model performance across scenarios\cite{Rothenberger2025}. \\ 
\hline

Combines complementary methods to overcome individual weaknesses. & 
Hybrid Approaches for Prediction Enhancement & 
Combining statistical methods and machine learning to address limitations of each method individually\cite{Yan2024}. & 
Improves accuracy and reduces uncertainty in predictions for complex and sensitive applications\cite{Yan2024}. \\ 
\hline

\end{longtable}

\textbf{Table \ref{table:methods-for-UQ-LLM}} encapsulates a variety of cutting-edge methods and applications designed to address the multifaceted challenge of uncertainty in medical LLMs. Each section highlights unique contributions, ranging from probabilistic learning techniques like Bayesian Neural Networks to hybrid approaches that combine machine learning and statistical methods. These methods target specific challenges, such as reducing noise, enhancing interpretability, and managing high-dimensional data. Notably, the use of frameworks like Monte Carlo Dropout and Dirichlet-multinomial models demonstrates the field's commitment to quantifying both epistemic and aleatoric uncertainties. Thematic approaches, such as establishing clinical vocabularies, not only refine prediction accuracy but also ensure that LLMs remain contextually relevant in dynamic medical environments.\\
The table also underscores the importance of human-centered elements in uncertainty management. Techniques like Grad-CAM and LIME facilitate interpretability, enabling clinicians to better understand model outputs and fostering trust in decision-making processes. High-uncertainty cases benefit from referral mechanisms, demonstrating the necessity of combining AI with human expertise in critical applications like ICU monitoring and clinical diagnostics. Additionally, frameworks such as GRADE and variance analysis provide structured mechanisms for evaluating performance under uncertainty, ensuring robust and comparable results across diverse scenarios. These advancements collectively highlight how integrating technical and human-centric approaches can significantly elevate the reliability and transparency of medical LLMs.\\

\subsubsection{Uncertainty Estimation}
In the context of LLMs for medicine, a diverse range of uncertainty estimation methods is employed to understand and mitigate ambiguity in the model's predictions. Techniques such as entropy-based measures, which compute predictive and semantic entropy, help quantify both types of uncertainty \cite{Xiao2022}\cite{Kucukgoz2025}. These methods are crucial for identifying unreliable outputs in high-stakes medical applications, where even minor errors can have significant implications.
Bayesian approaches—such as Maximum a Posteriori estimation and Gaussian Process emulators—have been adapted for LLMs to provide robust confidence measurements. These probabilistic techniques enhance the generalization of models by effectively incorporating uncertainty into the decision-making process, which is especially important when dealing with limited or noisy data in clinical scenarios \cite{Jones2025}\cite{Salvador2023}.
Advanced strategies, including deep ensembles and Monte Carlo (MC) dropout, as stated, are also used to further refine uncertainty quantification. Deep ensembles operate by training multiple LLMs independently and aggregating their predictions, thereby capturing the variability in outcomes. Conversely, MC dropout introduces stochasticity during inference, allowing the model to generate multiple outputs from which uncertainty can be estimated \cite{lakshminarayanan2017simple}\cite{ovadia2019can}. These approaches collectively facilitate a more comprehensive assessment of both types of uncertainty.
For text-based applications, such as clinical decision support or diagnostic report generation, token-level probability distributions and sample consistency methods have been developed to estimate uncertainty at a granular level. By analyzing the variability of generated tokens across multiple outputs, these techniques help detect potential hallucinations and ambiguous interpretations. However, the computational demands of such methods may limit their real-time applicability, necessitating careful calibration for specific tasks \cite{Liebermann2024}\cite{Koume2025}.
Visualization tools—such as uncertainty heatmaps or composite confidence metrics—play an essential role in translating raw uncertainty estimates into actionable insights for clinicians. These tools empower healthcare professionals to identify areas of low confidence and make more informed decisions. Additionally, advanced probabilistic frameworks, like Monte Carlo-Adaptive Sampling Mechanisms (MC-ASM), have shown promise in precisely estimating uncertainty, particularly when dealing with sparse or imbalanced datasets typical of specialized medical applications\cite{Huang2025}\cite{lanfredi2024enhancingchestxraydatasets}.
In summary, adapting these uncertainty estimation techniques for LLMs in medicine is critical to ensuring that the outputs are both reliable and interpretable. Such robust methods are integral to enhancing safety, building user trust, and ultimately improving clinical decision-making in environments where accurate, uncertainty-aware information is paramount.

\subsubsection{Mitigating Uncertainty}
Effective strategies were employed to mitigate uncertainty in medical applications by addressing its components both individually and collectively.Techniques such as Monte Carlo Dropout,deep ensembles,and Bayesian frameworks successfully reduced variability in predictions,offering robust methods for uncertainty quantification\cite{Wickstrøm2021}\cite{Molchanova2025}
\cite{Kucukgoz2025}.Enhancing data quality through preprocessing and augmentation also played a crucial role in mitigating input-related uncertainty,ensuring more reliable outputs in complex medical scenarios\cite{shorten2019survey}.
Thresholding mechanisms,structured explanations, and confidence scoring systems were introduced to minimize ambiguity and foster user trust, particularly in time-sensitive medical contexts where decisions have high stakes \cite{Bhutto2024}\cite{Campos2023}\cite{Huang2025}.User-centric interfaces designed with clear explanations enabled healthcare professionals to make informed decisions despite inherent uncertainties, aligning model outputs with clinical expectations\cite{ehsan2021expanding}.
Domain-specific adaptations, such as the incorporation of anatomical priors and multi-scale analyses, further enhanced the robustness of AI models under diverse clinical conditions. These adaptations allowed models to handle the variability often encountered in medical datasets, improving both performance and reliability\cite{Salvador2023}\cite{Dimitriou2023}\cite{Khawaled2024}\cite{Haq2025}.Advanced methods, including hybrid architectures and probabilistic loss functions, provided additional layers of precision and reliability, addressing both aleatoric and epistemic uncertainties in medical contexts\cite{Chen2025}.
Furthermore, integrating contextual information and adaptive models into AI systems addressed uncertainty stemming from environmental or situational factors. These approaches allowed models to dynamically adjust to changing conditions, enhancing relevance and robustness in real-world healthcare applications \cite{camacho2019four}. Combining technical robustness with user-centric design and ethical considerations ensured that AI systems not only met performance benchmarks but also aligned with societal values and clinical needs, making them more trustworthy and effective\cite{jobin2019global}\cite{floridi2019}.

\section{Proposed Thematic Framework: Summary of Methodological Approaches}
This research introduces a comprehensive framework designed to enhance uncertainty quantification in LLMs for medical applications. By integrating advanced probabilistic techniques with linguistic analysis, the framework offers a multidimensional approach that effectively addresses both epistemic and aleatoric uncertainties while ensuring model outputs are transparent and interpretable for clinical use.\\
Key Components of the Framework include:
\begin{itemize}
    \item \textbf{Probabilistic Modeling and Bayesian Inference:} \\
    The framework employs Bayesian methods—including Maximum a Posteriori estimation and Markov Chain Monte Carlo (MCMC) simulations—to capture the inherent probabilistic nature of model parameters and derive robust uncertainty distributions.
    
    \item \textbf{Hybrid Uncertainty Reduction Techniques:} \\
    It integrates advanced strategies such as deep ensembles and Monte Carlo dropout to generate multiple outputs. These outputs are used to compute entropy-based metrics and identify variability in predictions, thereby enabling a more comprehensive assessment of both epistemic and aleatoric uncertainty.
    
    \item \textbf{Linguistic Confidence Estimations:} \\
    In addition to numerical measures, the framework incorporates linguistic analysis by computing predictive and semantic entropy from the generated text. Sample consistency methods are utilized to gauge output stability across multiple iterations, enhancing interpretability.
    
    \item \textbf{Surrogate Modeling for Proprietary Systems:} \\
    To overcome the limitations of proprietary APIs (e.g., hidden internal probabilities in models like GPT-4), the framework integrates surrogate models such as Llama-2. These models provide access to internal probability distributions and improve overall model calibration.
    
    \item \textbf{Multi-Source Data Integration:} \\
    The framework supports the assimilation of both structured data (e.g., electronic health records, diagnostic imaging reports) and unstructured data (e.g., clinical notes, research articles). Advanced data fusion algorithms are employed to enhance input quality and mitigate uncertainties arising from domain shifts.
    
    \item \textbf{Dynamic Calibration and Adaptive Learning:} \\
    Continuous feedback mechanisms, including human-in-the-loop strategies, are integrated to enable real-time recalibration of the model. Techniques such as continual learning and meta-learning facilitate the model’s adaptability to evolving clinical contexts.
    
    \item \textbf{Explainability and Visualization Tools:} \\
    To bridge the gap between complex statistical outputs and clinical decision-making, the framework incorporates explainable AI methods. Uncertainty maps, trust scores, and composite confidence metrics are developed to provide clear, interpretable visualizations of uncertainty, thereby enhancing clinician trust.
    
    \item \textbf{Clinical Integration and Decision Support:} \\
    Finally, the framework aligns uncertainty metrics with clinical risk factors, ensuring that predictive outputs are contextually relevant. This integration supports decision-making systems by routing high-uncertainty cases for human expert review, ultimately fostering safer and more effective clinical outcomes.
\end{itemize}

\begin{figure}[h]
\centering
\hspace*{-1.5cm} 
\begin{tikzpicture}[
  node distance=1.1cm and 1.2cm, 
  every node/.style={align=center, font=\scriptsize}, 
  block/.style={rectangle, draw, fill=blue!20, rounded corners, minimum height=1cm, text width=2.2cm}, 
  arrow/.style={-{Latex[length=3mm,width=2mm]}, thick},
  dashedarrow/.style={dash pattern=on 3pt off 2pt, -{Latex[length=3mm,width=2mm]}, thick}
  ]

\node[block] (prob) {Probabilistic Modeling\\ \& Bayesian Inference};

\node[block, below left=of prob, xshift=-0.8cm] (hybrid) {Hybrid Uncertainty\\ Reduction Techniques};
\node[block, below right=of prob, xshift=0.8cm] (linguistic) {Linguistic Confidence\\ Estimations};

\node[block, below=of hybrid] (surrogate) {Surrogate Modeling\\ for Proprietary Systems};
\node[block, below=of linguistic] (multisource) {Multi-Source Data\\ Integration};

\node[block, below left=of surrogate, xshift=-0.4cm] (dynamic) {Dynamic Calibration\\ \& Adaptive Learning};
\node[block, below right=of multisource, xshift=0.4cm] (explain) {Explainability \&\\ Visualization Tools};

\node[block, below=2.3cm of $(dynamic)!0.5!(explain)$] (clinical) {Clinical Integration\\ \& Decision Support};

\draw[arrow] (prob) -- (hybrid);
\draw[arrow] (prob) -- (linguistic);

\draw[arrow] (hybrid) -- (surrogate);
\draw[arrow] (linguistic) -- (multisource);

\draw[arrow] (surrogate) -- (dynamic);
\draw[arrow] (multisource) -- (explain);

\draw[arrow] (dynamic) -- (clinical);
\draw[arrow] (explain) -- (clinical);

\draw[dashedarrow] (prob) -- (clinical);

\end{tikzpicture}
\caption{\textbf{Proposed Comprehensive Framework for Uncertainty Quantification in Medical LLMs.} The diagram illustrates an integrated structure starting from probabilistic modeling and Bayesian inference, which feeds into both hybrid uncertainty reduction techniques and linguistic confidence estimations. These components further lead to surrogate modeling for proprietary systems and multi-source data integration, respectively. Subsequent layers involve dynamic calibration with adaptive learning and explainability with visualization tools, which together converge into clinical integration and decision support. The dashed arrow emphasizes the overarching impact of probabilistic modeling on clinical outcomes.}
\label{fig:evolution-of-uq}
\end{figure}
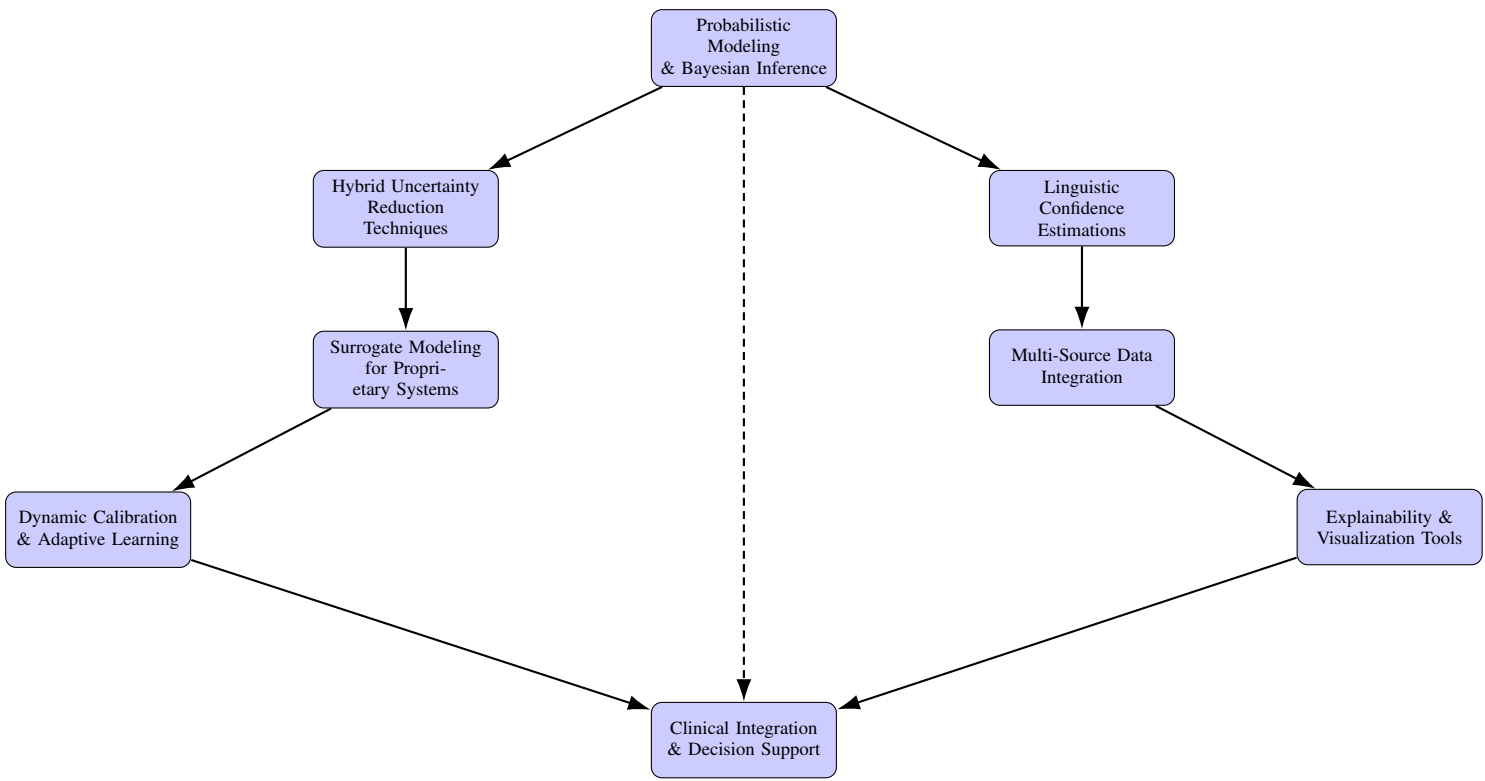
As shown in fig 7, our proposed framework of uncertainty quantification in medical LLMs, begins with probabilistic modeling and Bayesian inference as the foundational component. From this top-level node, the framework bifurcates into two primary branches: one focusing on hybrid uncertainty reduction techniques and the other on linguistic confidence estimations. Arrows indicate that the uncertainty captured through probabilistic modeling feeds into both branches, ensuring that foundational uncertainty measurements inform subsequent processes. The branch on hybrid techniques further develops into surrogate modeling for proprietary systems, while the linguistic branch leads to multi-source data integration. Each of these nodes contributes distinct uncertainty assessments—hybrid methods refine the numerical estimates using ensemble and dropout strategies, and linguistic methods extract semantic entropy and consistency measures from text outputs.\\
These processes eventually converge into advanced modules represented by dynamic calibration with adaptive learning and explainability with visualization tools. Arrows guide these two nodes downward, merging into clinical integration and decision support, which is designed to align and contextualize the uncertainty metrics with real-world clinical risk factors. The dashed arrow spanning directly from the top-level node to the final clinical integration module underscores the overarching influence of robust probabilistic modeling on clinical outcomes. By framing this integrated flow within the broader context of Responsible AI—and more specifically, Perceptual AI—this framework not only addresses the technical challenges of uncertainty quantification but also ensures that each stage promotes transparency, accountability, and user-centered design. In doing so, it aligns with the principles of Responsible AI by emphasizing interpretability, robustness, and ethical considerations, ultimately yielding AI systems that are not only accurate and reliable but also perceptually aware of their own limitations and the impact these may have on high-stakes clinical decision-making.\\

\section{Discussion}
The present research ventures into a nuanced exploration of uncertainty within LLMs for medical applications, situating its insights at the crossroads of technical rigor and philosophical inquiry. Rather than offering definitive resolutions, our approach proposes a dynamic, iterative process for managing uncertainty—a process that reflects the inherently provisional nature of knowledge itself.\\
Our work suggests that uncertainty, in its multifaceted forms, is not merely a technical challenge but a fundamental aspect of the epistemic landscape in which medical AI operates. Drawing upon themes in epistemology and the emerging discourse on Responsible and Perceptual AI, we posit that uncertainty can be reconceived as a reflective tool rather than as a limitation. By embracing uncertainty, our framework invites a more responsible deployment of AI, one that values transparency and user-centered design.\\
This perspective shifts the discourse from seeking absolute predictability to cultivating systems that are perceptually aware of their own limitations. Such systems, we argue, are better aligned with the ethical imperatives of clinical practice—a setting where decisions must account for both the known and the inherently unknowable.\\
From a Reflective AI perspective, this approach aligns with the idea that AI systems should not only process information but also critically assess their own reasoning and outputs. By embedding self-reflective mechanisms, AI can iteratively evaluate the reliability of its decisions and refine its explanatory capacity in response to user feedback. This paradigm supports the development of AI that is not just explainable but also dynamically attuned to the complexities of human-AI interaction, particularly in high-stakes domains like medicine.\\
In this sense, our contribution lies not in presenting final answers but in provoking deeper reflection on how uncertainty itself can inform more humane, accountable, and philosophically grounded AI systems in medicine \cite{lewis2024}.

\section{Conclusion}
This study has advanced our understanding of uncertainty quantification in LLMs within medical contexts by framing it as both a technical and philosophical challenge. Rather than pursuing absolute certainty, our approach embraces the inherent provisionality of knowledge—a recognition that uncertainty is not merely an obstacle but also a reflective tool that invites deeper inquiry into the limits of AI-mediated clinical decision-making.
By integrating advanced probabilistic methods with linguistic analysis and dynamic calibration, our framework underscores the necessity of a Responsible AI paradigm. This paradigm does not shy away from uncertainty; instead, it cultivates systems that are perceptually aware of their own limitations and are designed to operate transparently, ethically, and effectively in high-stakes environments. In a field as critical as healthcare, such a philosophical shift—from striving for complete predictability to accepting controlled, interpretable ambiguity—ensures that AI systems remain accountable and aligned with human values.
Ultimately, this research invites us to reconsider the role of uncertainty not merely as an engineering challenge but as a fundamental aspect of epistemology in AI. In accepting uncertainty as an intrinsic dimension of medical AI, could embracing controlled ambiguity fundamentally transform our understanding of what it means to truly "know" and "trust" in the context of clinical decision-making?

\section{ACKNOWLEDGEMENT} 
This research was undertaken, in part, thanks to funding from the Canada Research Chairs Program.

\bibliographystyle{unsrt}
\bibliography{references}

\section{appendix}
\begin{figure}[H]
   \centering
   \includegraphics[width=0.8\textwidth]{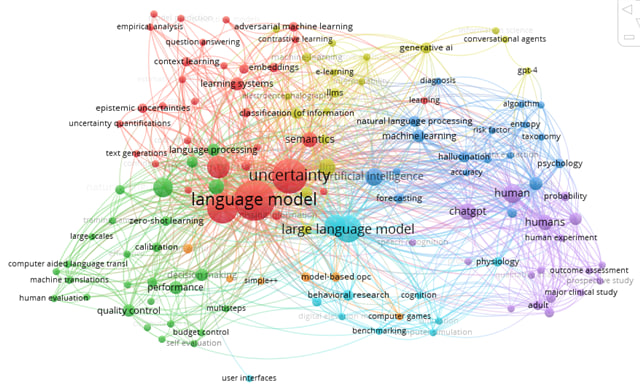} 
   \caption{Network Visualization\\96 Articles\\"Uncertainty" \& "LLM"\\Source: Scopus\\Keywords}
\end{figure}

\begin{figure}[H]
    \centering
    \includegraphics[width=0.8\textwidth]{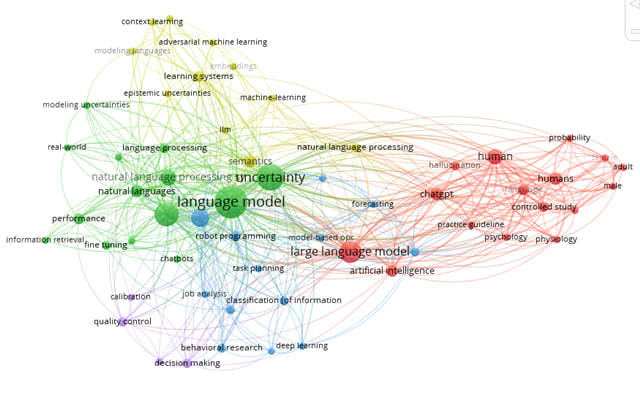} Assuming the image is named 2.jpg
    \caption{Network Visualization\\96 Articles\\"Uncertainty" \& "LLM"\\Source: Scopus\\Index Word}
\end{figure}
\begin{figure}[H]
    \centering
    \includegraphics[width=0.8\textwidth]{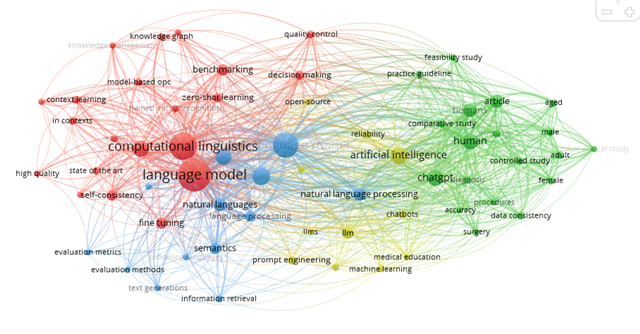} 
    \caption{Network Visualization\\194 Articles\\"Consistency" \& "LLM"\\Source: Scopus\\Keywords}
\end{figure}

\begin{figure}[H]
\centering
    \includegraphics[width=0.8\textwidth]{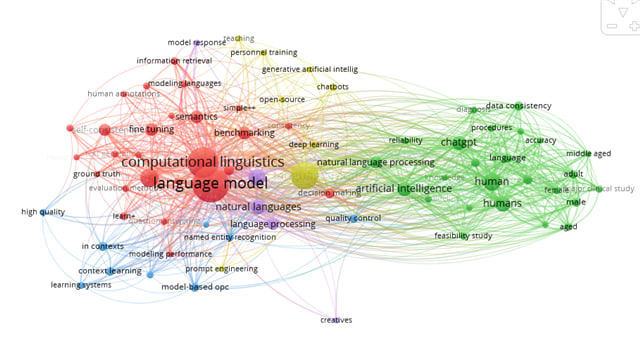} 
    Assuming the image is named 4.jpg
    \caption{Network Visualization\\194 Articles\\"Consistency" \& "LLM"\\Source: Scopus\\Index Word}
\end{figure}

\end{document}